%% file: neurips_2023.tex
\documentclass{article}

\usepackage[preprint]{neurips_2023}

\usepackage[utf8]{inputenc} 
\usepackage[T1]{fontenc}    
\usepackage[hidelinks]{hyperref}       
\usepackage{url}            
\usepackage{booktabs}       
\usepackage{amsfonts}       
\usepackage{nicefrac}       
\usepackage{microtype}      
\usepackage{xcolor}         
\usepackage{siunitx}
\usepackage{graphicx}
\usepackage{arydshln}
\usepackage{wrapfig}
\usepackage{enumitem}
\usepackage{multirow}
\usepackage{xspace}

\newcommand{\modelname}{\textsc{T\"ulu}\xspace}
\newcommand{\codemodelname}{\textsc{Code T\"ulu}}

\newcommand{\llama}{\textsc{Llama}}
\newcommand{\codellama}{\textsc{Code Llama}}
\newcommand{\mistral}{\textsc{Mistral}}
\newcommand{\instructgpt}[1]{\text{Davinci}-{\text{#1}}}

\definecolor{darkred}{RGB}{156, 39, 33}
\definecolor{darkblue}{RGB}{31, 90, 153}

\title{Camels in a Changing Climate:\\Enhancing LM Adaptation with \modelname~2}

\newcommand*\samethanks[1][\value{footnote}]{\footnotemark[#1]}
\author{
    Hamish Ivison\thanks{Equal contribution.} ~$^{\spadesuit}$ \; \; 
    Yizhong Wang\samethanks{} ~$^{\clubsuit\spadesuit}$ \; \; 
    \textbf{Valentina Pyatkin$^{\clubsuit\spadesuit}$ \; \; 
    Nathan Lambert$^\clubsuit$} \; \; \\
    \textbf{Matthew Peters$^\clubsuit$ \; \;
    Pradeep Dasigi$^\clubsuit$ \; \;} 
    \textbf{
    Joel Jang~$^{\clubsuit\spadesuit}$ \; \;
    David Wadden$^\clubsuit$} \; \; \; \\
    \textbf{Noah A. Smith$^{\clubsuit\spadesuit}$ \; \; \; 
    Iz Beltagy$^\clubsuit$ \; \; \; 
    Hannaneh Hajishirzi$^{\clubsuit\spadesuit}$} \\ \\
    $^\clubsuit$Allen Institute for AI \;
    $^\spadesuit$University of Washington \\
    \texttt{\{yizhongw,hamishiv\}@cs.washington.edu}
}

\begin{document}

\maketitle

\begin{abstract}



  Since the release of \modelname{}~\citep{wang2023far}, open resources for instruction tuning have developed quickly, from better base models to new finetuning techniques. We test and incorporate a number of these advances into \modelname{}, resulting in \modelname~2, a suite of improved \modelname models for advancing the understanding and best practices of adapting pretrained language models to downstream tasks and user preferences. Concretely, we release: (1) \textbf{\modelname-V2-mix}, an improved collection of high-quality instruction datasets; (2) \textbf{\modelname 2}, \llama{}-2 models finetuned on the V2 mixture; (3) \textbf{\modelname 2+DPO}, \modelname 2 models trained with direct preference optimization (DPO), including the largest DPO-trained model to date (\textbf{\modelname 2+DPO 70B}); (4) \textbf{\codemodelname{} 2}, \codellama{} models finetuned on our V2 mix that outperform \codellama{} and its instruction-tuned variant, \codellama-Instruct. Our evaluation from multiple perspectives shows that the \modelname{}~2 suite achieves state-of-the-art performance among open models and matches or exceeds the performance of GPT-3.5-turbo-0301 on several benchmarks. We release all the checkpoints, data, training and evaluation code to facilitate future open efforts on adapting large language models.
  

\end{abstract}

\section{Introduction}
The capabilities of large language models (LMs) to follow user requests have been progressing rapidly through a wide range of openly available models, datasets, and training methods.
Since the release of the original \modelname{} models~\citep{wang2023far}, there have been a number of significant advances in almost all aspects of  language model adaptation, from the release of improved finetuning datasets~\citep{UltraChat, cui2023ultrafeedback}, to  increasingly powerful base models~\citep{touvron2023llama,jiang2023mistral}, to powerful and accessible adaptation methods for combining these components~\citep{rafailov2023direct, dettmers2023qlora}.
We comprehensively evaluate and combine these recent advances to present strong open models across 7, 13, and 70 billion parameter scales with empirical studies of various training recipes.

Accompanying our new models, we release a new dataset mixture, \textbf{\modelname-V2-mix} that results in stronger performance across a variety of reasoning and knowledge-probing tasks. We also compare the performance of both new parameter efficient tuning and reinforcement learning from human feedback (RLHF) methods.
Included in our model suite is a \llama{}-2 70B model finetuned on \modelname-V2-mix and further trained using direct preference optimization (DPO) algorithm, representing \textbf{the first stable demonstration of using DPO at scales of 70 billion parameters}. This model achieves results competitive with state-of-the-art on the MT-Bench and AlpacaEval benchmarks.

We additionally explore training with quantized low-rank adaptation (QLoRA), finding that it solid performance across traditional language processing tasks, but falls behind on evaluations that examine long-form text generation such as AlpacaEval.
Finally, we apply our mixture to \codellama~\citep{roziere2023code}, resulting in \codemodelname{} 2, which outperforms both the base \codellama model and its instruction-tuned variant \codellama-Instruct across all model sizes.

\modelname{}-2 validates and extends the progress seen across many open instruction model recipes released recently, such as those with some RL component, including Zephyr-Beta~\citep{tunstall2023zephyr}, \llama-2-chat~\citep{touvron2023llama}, XWin~\citep{xwin-lm}, WizardLM~\citep{xu2023wizardlm}, and OpenChat~\citep{wang2023openchat}, and some without, including \mistral-Instruct~\citep{jiang2023mistral} and Mosaic Pretrained Transformer (MPT)~\citep{mptllama}.

In summary, with \modelname{}~2, we find that:
\begin{enumerate}
    \item Recent \textbf{distilled data mixtures have significantly improved} in terms of downstream performance over both instruction and preference datasets available only six months ago, with our new mixture outperforming our old mixture by an average of 8\%.
    \item \textbf{DPO training scales to 70 billion parameter models}, and significantly improves open-ended generation metrics without degrading model capabilities, improving AlpacaEval performance by an average of 13\% across model sizes. Our largest DPO trained model, \modelname~2+DPO 70B, achieves state-of-the-art performance for MT-Bench~\citep{zheng2023judging} compared to open-weight models.
    \item \textbf{QLoRA training does not match full-finetuning in long-form generation tasks}, although the gap shrinks with model size (from 10\% worse on average to 3\% worse on average across our tasks). We note that QLoRA especially underperforms on open-ended generation tasks such as AlpacaEval (20\% average gap in performance).
    \item \textbf{\codemodelname{} 2 significantly improves coding abilities over \modelname~2} (70\% average improvement in Codex-Eval) but degrades open-ended model generations in AlpacaEval (20\% average drop in performance).
\end{enumerate}

We publicly release all models, data, and code associated with this work.
Models and the new dataset mix can be found at \url{https://huggingface.co/collections/allenai/tulu-v2-suite-6551b56e743e6349aab45101}.
Our finetuning and evaluation code can be found at \url{https://github.com/allenai/open-instruct}. 
We hope that publicly releasing all artifacts aids future research into post-pretraining LM adaptation.

\section{\modelname{} V2 Details}

We first detail the aspects of adaptation we explored for \modelname~2 in comparison to \modelname~1~\citep{wang2023far}: new base models, a new data mixture, extended context training data, and RLHF training.
\modelname~1 constructed two data instruction mixes through a variety of experiments, one containing prompt-response pairs fully written by humans from the FLAN, Dolly and Open Assistant datasets, and another containing prompt-response pairs fully or partially generated by OpenAI models along with the human-written data.

\paragraph{Improved base models}

We first switch from using \textsc{LLaMa}-1 models~\citep{touvron2023llama} to \llama{}-2~\citep{touvron2023llama2}, a newer set of models following similar architecture to \textsc{LLaMa}-1 but pretrained on significantly more tokens (2 trillion tokens as opposed to 1 or 1.4 trillion tokens), and displaying improved performance (\citet{touvron2023llama2} shows a 10\% average improvement across model sizes on a set of academic benchmarks).
We also experiment with \codellama, a set of \llama{}-2 models further pretrained on code data. 
We finetune models at all possible \llama{}-2 sizes: 7B, 13B, and 70B, and all possible \codellama{} sizes: 7B, 13B, and 34B.

\paragraph{V2 data mixture}

Our original data mixture (\modelname-V1-mix) was based on ablations over human and GPT-generated datasets -- we refer readers to~\citet{wang2023far} for a full list. We keep a number of high-quality datasets from our first mix, and add new datasets that are either carefully manually curated for quality or generated from GPT models while encouraging complexity and diversity. We additionally downsample larger datasets such as FLAN to reduce the overall size of the training mixture, and remove Dolly~\citep{dolly} from the mixture due to its poor performance in previous ablations. Our V2 mixture, \modelname-V2-mix, comprises of data from the following sources (we mark  datasets newly added to our V2 mixture with *):

\begin{itemize}[leftmargin=*]
    \item \textbf{FLAN}~\citep{flant5}: We use 50,000 examples sampled from FLAN v2.
    \item \textbf{CoT}: To emphasize chain-of-thought (CoT) reasoning, we sample another 50,000 examples from the CoT subset of the FLAN v2 mixture.
    \item \textbf{Open Assistant 1}~\citep{kopf2023openassistant}: We isolate the highest-scoring paths in each conversation tree and use these samples, resulting in 7,708 examples. Scores are taken from the quality labels provided by the original annotators of Open Assistant 1.
    \item \textbf{ShareGPT}\footnote{\label{vicuna_note}
ShareGPT (\url{https://sharegpt.com/}) data was used to build the Vicuna model \citep{vicuna2023}, but the exact dataset has not been released. Following~\citet{wang2023far}, we instead use a reproduced version from \url{https://huggingface.co/datasets/anon8231489123/ShareGPT_Vicuna_unfiltered/tree/main/HTML_cleaned_raw_dataset}, and follow Vicuna to split the long conversations into blocks with a maximum length of 4,196 tokens.}: We use all 114,046 examples from our processed ShareGPT dataset, as we found including the ShareGPT dataset resulted in strong performance in prior work.
    \item \textbf{GPT4-Alpaca}~\citep{peng2023instruction}: We sample 20,000 samples from GPT-4 Alpaca to further include distilled GPT-4 data.
    \item \textbf{Code-Alpaca}~\citep{codealpaca}: We use all 20,022 examples from Code Alpaca, following our prior V1 mixture, in order to improve model coding abilities.
    \item \textbf{*LIMA}~\citep{zhou2023lima}: We use 1,030 examples from LIMA as a source of carefully curated data.
    \item \textbf{*WizardLM Evol-Instruct V2}~\citep{xu2023wizardlm}: We sample 30,000 examples from WizardLM, which contains distilled data of increasing diversity and complexity.
    \item \textbf{*Open-Orca}~\citep{OpenOrca}: We sample 30,000 examples generated by GPT-4 from OpenOrca, a reproduction of Orca~\citep{mukherjee2023orca}, which augments FLAN data with additional model-generated explanations.
    \item \textbf{*Science literature}: We include 7,544 examples from a mixture of scientific document understanding tasks--- including question answering, fact-checking, summarization, and information extraction. A breakdown of tasks is given in Appendix~\ref{app:science_mix}.
    \item \textbf{*Hardcoded}: We include a collection of 140 samples using prompts such as `Tell me about yourself' manually written by the authors, such that the model generates correct outputs given inquiries about its name or developers.
\end{itemize}

Additionally, we filter any samples that include references to other LLM systems such as GPT-4, Open Assistant, or Claude, to avoid contradicting the hardcoded prompts. After filtering, the V2 mixture consists of 326,154 samples, compared to 490,445 in the V1 mixture. Our dataset is available at \url{https://huggingface.co/datasets/allenai/tulu-v2-sft-mixture}.

\paragraph{Extended context length}

\begin{figure}
    \centering
    \vspace{-10pt}
    \includegraphics[width=.5\textwidth, trim=0.6cm 0cm 0cm 0cm]{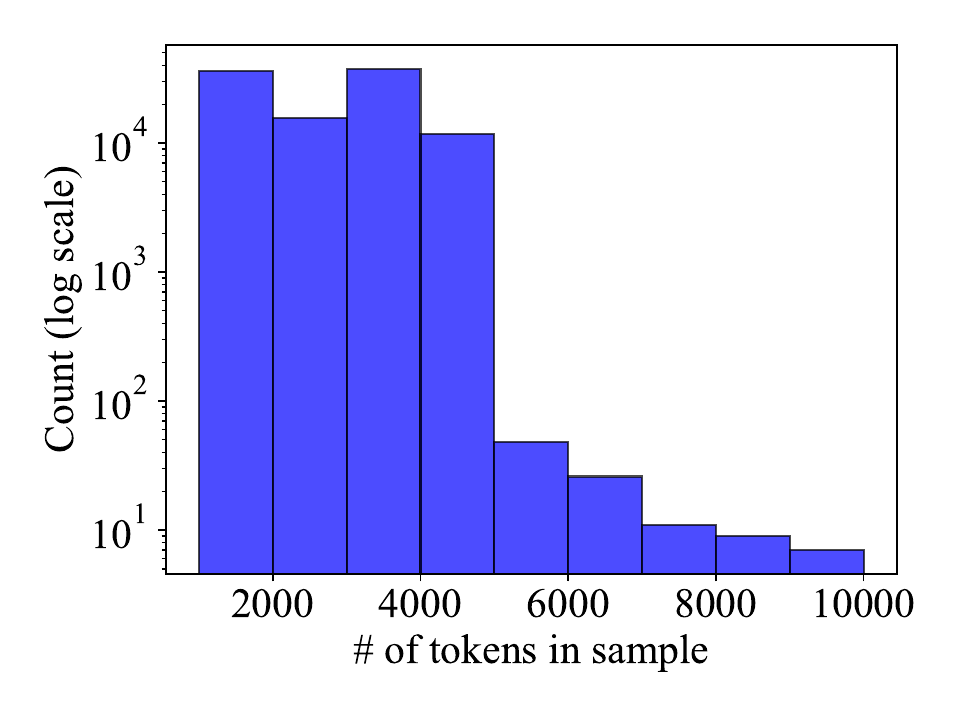}
    \caption{Histogram of token lengths in our V2 data mixture.}
    \label{fig:length_histogram}
\end{figure}

We expand the context length during training from a maximum of 2,048 tokens to 8,192 tokens in order to make better use of the many lengthy samples in datasets such as ShareGPT and Open Assistant 1. 
Moving from 2,048 to 8,192 max length means we only truncate 20 (as opposed to 63,900) samples within our V2 mixture, better capturing the long tail of lengthy examples in our training data. 
We plot the length distribution of our V2 mixture in Figure~\ref{fig:length_histogram}. The mean length of a sample is 1097 tokens, with the 25th and 75th percentile values being 230 and 1464 respectively.

\paragraph{RLHF training}

Reinforcement learning from human feedback (RLHF) is a core component of modern user-facing LLM systems~\citep{bai2022training, ouyang2022training, touvron2023llama}.
Early systems for RLHF were built primarily upon the proximal policy optimization (PPO) algorithm, but recent advances have seen exploration of offline RL~\citep{snell2022offline}, reward model data filtering called rejection sampling (RS)~\citep{touvron2023llama} or reinforced self-training (ReST)~\citep{gulcehre2023reinforced} and direct integration of preference data~\citep{rafailov2023direct}.
In this work, we use the direct preference optimization (DPO) algorithm due to the simplicity of its implementation~\citep{rafailov2023direct}. 
For DPO training, we follow the Zephyr-Beta approach~\citep{tunstall2023zephyr}: we train on a filtered and binarized form of UltraFeedback~\citep{cui2023ultrafeedback} for three epochs. 
One thing to note is the low learning rate, \num{5e-7}, required for stable and effective DPO training.
We find this significantly improves performance on open-ended generation evaluations such as AlpacaEval~\citep{alpaca_eval}, while making little to no difference in performance over more capability-focussed evaluations such as MMLU and HumanEval.

\paragraph{QLoRA training}

We experimented with QLoRA training at the instruction tuning stage in order to determine if we could reduce our compute demands without reducing performance. Due to sub-par performance at the instruction tuning stage, we did not explore using QLoRA during RLHF training, although we note that prior work has found it to perform well for PPO-based RLHF training~\citep{santacroce2023efficient, sun2023exploring}.

\section{Experiments}

\paragraph{Evaluation tools}
We reuse the evaluation framework from \modelname{} 1~\citep{wang2023far}, which includes evaluations testing factual knowledge (MMLU), reasoning (GSM8k, Big Bench Hard), multilinguality (TydiQA), coding (CodexEval), open-ended generation (AlpacaEval), toxicity (ToxiGen), and truthfulness (TruthfulQA). 
We refer the reader to \citet{wang2023far} for a more in-depth explanation of these evaluations, and provide an overview of each evaluation in Appendix~\ref{app:evaluation_suite}.

We make two changes to this evaluation framework: first, we replace our old AlpacaFarm setup with the default AlpacaEval setup~\citep{alpaca_eval}, making our reported numbers directly comparable with the AlpacaEval leaderboard (\url{https://tatsu-lab.github.io/alpaca_eval/}). At time of writing, AlpacaEval does not use a pinned GPT-4 version for evaluation, so we ensure all evaluations reported use GPT-4-0613 as the evaluator model. Second, we also evaluate a set of models on MT-Bench~\citep{zheng2023judging}, a popular benchmark for open-ended generation that similarly uses GPT-4 to judge model outputs across a diverse set of prompts.

While TruthfulQA is included in our evaluation suite, we found that the data used for DPO training (UltraFeedback) made use of TruthfulQA prompts. 
As such, we omit TruthfulQA results when showing comparisons with contaminated models (any models trained with the UltraFeedback dataset). 
We also note that although we report results for several GPT models (GPT-4-0314, GPT-3.5-turbo-0301, GPT-4-1106-preview), \textbf{we cannot rule out the possibility they are trained on the evaluation benchmark datasets}.

\paragraph{Training} We detail the hyperparameters used to train models in Appendix~\ref{app:hyperparams}. The 70B variant of \modelname{}~V2-DPO was trained on a 512-core TPUv3, completing three epochs in approximately 7 days. 

\subsection{Overall Results}

\input{tables/baseline_comparisons}

We present our overall results comparing \modelname-2 to popular proprietary and open models in Table~\ref{tab:baseline_comparisons}. We find that:

\paragraph{\modelname 2 outperforms all open models on average.} \modelname-2 70B is the highest-performing model on average and is the best-performing open model in 3/7 tasks. For the remaining 4 tasks, it is outperformed in MMLU and CodexEval by \modelname 2+DPO 70B, in ToxiGen by \llama{}-2-Chat models, and in AlpacaEval by Xwin-LM 70B. We note that the average gap between \modelname 2 70B and the highest performing model in these 4 tasks is under 1\%, highlighting that \modelname 2 is at least competitive if not outright better than all open models in most evaluations.

\paragraph{\modelname 2 is competitive with GPT 3.5-0301.} \modelname 2 70B achieves similar performance to GPT-3.5-turbo-0301 in MMLU, BBH and TydiQA, and outperforms it in AlpacaEval and ToxiGen. However, there remains a large gap with GPT-4 and a moderate gap with GPT-3.5-turbo-0613 (a more modern variant of the model) in most evaluations.

\paragraph{Scaling trends remain strong with \modelname 2.} Increasing model size improves almost every metric when the finetuning setup is held consistent across our model suite.



\subsection{\modelname V1 vs V2 Data Mixtures}

\input{tables/mix_comparison_table}

We compare our new model suite to our old models in Table~\ref{tab:mix_comparisons}, comparing \llama{}-2 models at all sizes on our V1 and V2 mix. We additionally compare our V2 mix to a model trained only on ShareGPT, the most promising single dataset from our original work. We find that:

\paragraph{Models trained on the V2 mix perform better than models trained on the V1 mix on open-ended generation.} V2 mix models outperform V1 mix models consistently on BBH, Codex-Eval, AlpacaEval, and TruthfulQA, and consistently underperform the V1 mix on GSM8k and TydiQA. The former is likely due to training on fewer CoT examples (which contains the GSM8k train dataset), while the latter indicates our V2 mix is worse for multilingual capabilities. This reinforces the findings from \citet{wang2023far} that no one dataset is optimal for all tasks, although we note on average models trained on our V2 mix outperform those trained on our V1 mix.

\paragraph{Models trained on the V2 mix outperform training on ShareGPT across most evals.} In prior work and in Table~\ref{tab:mix_comparisons}, we find that training on ShareGPT alone results in overall performance close to models trained on our V1 mix, and greatly improved AlpacaEval performance. However, our new mix actually outperforms using ShareGPT alone both overall and only considering AlpacaEval. This is likely due to the V2 mix's greater reliance on distilled datasets that have similar origins to ShareGPT.

\paragraph{Improvements from the V2 mix shrink with model size.} While the V2 mix provides a 13\% average improvement at the 7B scale, it only provides a 1\% improvement at the 70B scale. This suggests that the importance of instruction data quality may shrink as model size (and/or capabilities) increase.

Having established the overall superiority of our V2 mix, especially on open-ended generation, we now turn to alternate finetuning methods to further improve \modelname{} 2.

\subsection{Scaling DPO Training}

\input{tables/dpo_comparisons}

We finetune our models using DPO~\citep{rafailov2023direct} and the Ultrafeedback dataset~\citep{cui2023ultrafeedback}, following the hyperparameters and overall setup used by Zephyr-Beta~\citep{tunstall2023zephyr}, who apply DPO to a 7B Mistral model finetuned on UltraChat~\citep{UltraChat}. Surprisingly, we find these hyperparameters scale, providing stable training and performance improvements for models at all sizes. We show our results in Table~\ref{tab:dpo_comparisons} and results focusing on GPT-based evaluations (MT-Bench and AlpacaEval) in Table~\ref{tab:mtbench}. We provide full MT-Bench results in Appendix~\ref{app:full_mt_bench}. We find that:

\paragraph{DPO training significantly improves AlpacaEval and MT-Bench performance.} At all sizes, DPO training provides significant improvements in AlpacaEval, with our largest DPO-trained model significantly outperforming GPT-3.5-turbo-0314 (89.4 vs.~95.1) and is competitive with GPT-4 (see Table~\ref{tab:mtbench}. \textbf{\modelname 2+DPO 70B is the second best-performing open model on AlpacaEval},\footnote{At time of writing. See \url{https://tatsu-lab.github.io/alpaca_eval/}} just behind Xwin-LM 70B. We also observe that DPO training provides a large boost in MT-Bench performance for the 13B and 70B size models, with \textbf{\modelname{} 2+DPO 70B being the best-performing open model compared to all other models on the MT-Bench leaderboard}.\footnote{At time of writing. See \url{https://huggingface.co/spaces/lmsys/chatbot-arena-leaderboard}} Curiously, while \modelname~2 outperforms most GPT models we examine in AlpacaEval, it underperforms compared to all of them in MT-Bench.

\input{tables/mtbench}

\paragraph{DPO training is stable at large scales.} 
We find that DPO training scales without issues with 70B-size models, with DPO training still providing large benefits for open-ended generation (AlpacaEval) even at the 70B size.
This suggests DPO is a promising path for training large models on human feedback without the engineering complexity required by PPO.
To our knowledge, \modelname~2+DPO 70B is the largest publicly-released DPO-trained model.

\paragraph{DPO does not dramatically harm most other metrics.} 
We find that DPO training does not significantly change performance in most other metrics we measure, such as factual reasoning (MMLU) or reasoning (BBH, GSM8k), with the exception of multilinguality (which we discuss below). This suggests that DPO training does not significantly change model capabilities.

\paragraph{DPO training significantly drops multilingual capabilities.} We find that DPO training significantly drops performance in TydiQA, which tests the multilingual capabilities of our model. However, we note that both our supervised finetuning and DPO data mixes do not explicitly contain multilingual data, and are majority English-language. As such, DPO training is likely to make multilingual outputs further out-of-distribution, and mixing in multilingual data at instruction tuning and DPO training stages may significantly improve these results.

\paragraph{DPO training increases model verbosity.} As seen in Table~\ref{tab:mtbench}, \modelname~2+DPO models generally output answers of longer length than those trained without DPO. This is in line with prior work showing a bias toward verbosity from RLHF training~\citep{dubois2023alpacafarm,singhal2023long}. However, we note that our DPO-trained models appear dramatically less verbose than other open-weight models, which future work will investigate.

\subsection{Parameter-efficient Finetuning}

In order to reduce compute demands, we experimented with using quantized low-rank adaptation (QLoRA)~\citep{dettmers2023qlora} at the instruction tuning stage. We followed the suggested hyperparameters from \citet{dettmers2023qlora} and trained \llama{}-2 models at all sizes using QLoRA. We compare these to our fully-finetuned \modelname~2 models (without DPO) in Table~\ref{tab:qlora_comparisons}. We find:

 

\input{tables/qlora_comparisons}


\paragraph{QLoRA struggles on open-ended generation tasks.} We observe that QLoRA underperforms full-finetuning in AlpacaEval in a consistent manner, likely due to the open-ended nature of the task. We suggest the discrepancy of our results compared to \citet{dettmers2023qlora} may be due to the wider set of tasks in our evaluation suite, as \citet{dettmers2023qlora} focusses on MMLU performance as a way to compare QLoRA and full-finetuning performance (where we do see much closer performance between QLoRA and full-finetuning). In our overall average, we observe a gap between QLoRA and full-finetuning.

\paragraph{The gap between QLoRA and full-finetuning shrinks with size.} Similar to prior work in parameter-efficient learning~\citep{lester-etal-2021-power}, we find that the average gap in performance between QLoRA and full-finetuning shrinks with model size, suggesting that QLoRA may start to match full-finetuning at even larger model sizes.

\subsection{Improving Code Performance with \codellama}

\input{tables/codellama_comparisons}

Finally, we attempted using \codellama~\citep{roziere2023code} as a base model instead of \llama{}-2 due to its improved performance on coding tasks. We dub \codellama{} models trained on our V2 data mixture as \codemodelname{} 2 models. We present our results comparing \codellama{} and \llama{}-2 models fully finetuned on our V2 mixture in Table~\ref{tab:codellama_comparisons}. We find that:

\paragraph{\codemodelname~2 models significantly outperform \modelname 2 models at coding tasks.} As expected, \codemodelname~2 models report drastically improved Codex-Eval performance compared to \modelname~2 -- in Codex-Eval, our smallest (7B) \codemodelname{}~2 model matches the performance of \modelname{}-V2+DPO 70B, our strongest \llama{}-2-based model. This highlights the efficacy of using smaller, domain-specific models when limiting evaluation to that domain alone. 

\paragraph{\codemodelname{}~2 and \modelname~2 display drastically different results across non-code evaluations.} While we can only compare two sizes, we find that \modelname~2 models consistently outperform \codemodelname{}~2 models in 4 out of 8 tasks (MMLU, GSM8k, AlpacaEval, TruthfulQA), while \codemodelname{}~2 performs well in BBH, TydiQA, ToxiGen, and Codex-Eval. Since \codellama{} models are variants of \llama{}-2 models additionally pretrained on code data, this suggests the continued code pretraining has significantly altered model capabilities. In particular, we note that performance on AlpacaEval appears to drop by a large margin (by around 20\%).

\paragraph{Code \modelname~2 outperforms \codellama-base and \codellama-Instruct across all sizes.} We find that \codemodelname~2 models, using our V2 data mix, outperform both base \codellama{} and \codellama-Instruct models in 5 our of 8 evaluation settings (and are stronger on average), highlighting the efficacy of our V2 data mixture. \codellama-Instruct was finetuned on an internally developed private dataset we do not have access to, which makes it difficult to compare to our mixture, but the strong performance of \codellama-Instruct on AlpacaEval suggests the mixture may focus on general open-ended queries rather than specific model capabilities.

We release our \codemodelname{}~2 models alongside the rest of our V2 suite.

\section{Conclusion}

We present \modelname~2, a set of models, along with recipes for continuing the progress of fine-tuning LMs across a variety of tasks.
This release represents a strong incremental step through better performance of the new data mixture, stability of DPO training, and comparison to parameter-efficient training methods.

Substantial work is still needed to understand the mechanisms causing the improvement in performance from these datasets and the DPO training methodology. Future work could involve more investigation of the impact of methods such as DPO on handling refusal behaviour, investigating the impact of different data ablations on DPO performance, and performing comparisons to other RLHF algorithms (e.g., PPO) at scale. Additionally, incorporating improved base models will likely yield further gains over the models presented here. We hope such work can be enabled by the public release of all our data, code, and models. 

\section*{Acknowledgments}
Research supported by Cloud TPUs from Google's TPU Research Cloud (TRC). We thank Eric Mitchell and Rafael Rafailov for helpful discussions involving DPO training dynamics.

\bibliographystyle{abbrvnat}
\bibliography{anthology,custom}

\clearpage
\appendix

\section{Evaluation Suite}
\label{app:evaluation_suite}
We describe our evaluation suite below for easy reference:
\begin{itemize}[leftmargin=*]
    \item \textbf{MMLU}: We use the official MMLU evaluation script and prompts available at \url{https://github.com/hendrycks/test}, with modifications to allow for batch processing. We evaluate using 0 few-shot examples, following the original setup of MMLU. We report average accuracy across test examples.
    \item \textbf{GSM}: We evaluate models on the test set of GSM. Following \citet{wei2022chain}, we evaluate with chain-of-thought. We use 8 few-shot in-context examples. Because all answers in GSM are numbers, we extract the last number in the model response as the final answer. We report average accuracy across test examples.
    \item \textbf{BBH}: We follow the setup described in the original paper \citet{suzgun2022challenging}, and evaluate with chain-of-thought. The officially provided prompts, which have 3 few-shot in-context examples are used. For the CoT setup, we extract the first word after the phrase `So the answer is', or the entire response if there is no such substring present. We report average accuracy over sub-tasks (all of which use accuracy as the primary metric).
    \item \textbf{TydiQA}: We follow the setup described in the PaLM 2 technical report~\citep{anil2023palm} to evaluate models' performance in answering multilingual questions. We report only one setting, GP, where the gold passage that contains the answer is given (GoldP/GP). One in-context example is used to familiarize the model with the answering format.
    \item \textbf{Codex-Eval}: We use the HumanEval dataset in the Codex paper \citep{chen2021evaluating} for evaluating models' coding ability. The dataset contains 164 programming problems, where models are prompted to complete the Python function given its docstring. Following the original paper, we compute unbiased estimates of pass@k to measure the functional correctness of models' outputs. We report pass@10. We sample with a temperature of 0.8.
    \item \textbf{ToxiGen}: We follow the setup in \citet{touvron2023llama2}, but use the original set of prompts from \citet{hartvigsen2022toxigen}, which are designed to elicit toxic generations for certain groups. We take only the prompts designed to produce toxic language (`hateful' prompts) and use 500 prompts per group to reduce evaluation costs. For base language models, we pass in the original ToxiGen prompts unchanged and greedily decode up to the first new line (or a maximum of 512 tokens). For instruction-tuned models, we place the prompt in the corresponding template, and ask the model to complete the prompt, until the model generates a stop token (or a maximum of 512 tokens). We pass the generated text into a roberta-large model trained to detect toxic content finetuned as part of \citet{hartvigsen2022toxigen}\footnote{\href{https://huggingface.co/tomh/toxigen\_roberta}{https://huggingface.co/tomh/toxigen\_roberta}}. We then report the percentage of generations deemed toxic by the classifier.
    \item \textbf{TruthfulQA}: Following \citet{touvron2023llama2}, we mainly use the generation setting of TruthfulQA \citep{lin2022truthfulqa}. The TruthfulQA dataset contains 818 questions, which are used to prompt the tested model to generate answers. We use the default QA prompt format with 6 in-context QA examples. We follow the official script in their official implemention\footnote{https://github.com/sylinrl/TruthfulQA/} to do greedy decoding and answer postprocessing. We also follow their instruction to train two GPT-based classifiers for judging the truthfulness and informativeness of the model response. We report the rate of the responses being truthful and informative (\% Informative and Truthful) following \citet{touvron2023llama2}. We only report the \% Informative and Truthful as our primary metric.
    \item \textbf{AlpacaEval}: We use the package provided by \citet{alpaca_eval}, following the default setup which asks the evaluated model to generate responses for 805 prompts and employ GPT-4 to compare the response with \instructgpt{003}. We employ the ``alpaca\_eval\_gpt4'' annotator. We allow the evaluated model to generate up to 8192 tokens, without specifying special stop sequences. The reported win-rate is the percentage of model generations that GPT-4 reports as being preferred over the generations from \instructgpt{003}.
    \item \textbf{MT-Bench}: We use the single-answer grading setting of MT-Bench, as suggested by the MT-Bench repository\footnote{\href{https://github.com/lm-sys/FastChat/tree/main/fastchat/llm_judge\#mt-bench}{https://github.com/lm-sys/FastChat/tree/main/fastchat/llm\_judge\#mt-bench}}. MT-Bench consists of 80 questions with followups, resulting in 160 responses being graded by a GPT-4 model across varying domains. While MT-Bench does not have a pinned GPT-4 version, we ensure all reported evaluations use GPT-4-0613.
\end{itemize}

\section{Training Hyperparameters}
\label{app:hyperparams}

For instruction-tuning/supervised fine-tuning, our training hyperparameters were as follows:
\begin{itemize}
    \item Precision: BFloat16
    \item Epochs: 2
    \item Weight decay: 0
    \item Warmup ratio: 0.03
    \item Learning rate: 2e-5 (1e-5 for 70B)
    \item Max. seq. length: 8,192
    \item Effective batch size: 128
\end{itemize}

For QLoRA training, we used the following:
\begin{itemize}
    \item Epochs: 5
    \item Weight decay: 0
    \item Warmup ratio: 0.03
    \item Learning rate: 1e-4
    \item Max. seq. length: 4,096
    \item Effective batch size: 128
    \item LoRA Rank: 64
    \item LoRA Alpha: 16
    \item LoRA dropout: 0.1
    \item Layers wrapped: all attention and feedforward linear layers
\end{itemize}

We experimented with a variety of QLoRA hyperparameters and found in smaller-scale experiments that these were the best hyperparameters we could fit into our compute budget while still giving strong performance.

For DPO, we used the following hyperparameters:
\begin{itemize}
    \item Precision: BFloat16
    \item Epochs: 3
    \item Weight decay: 0
    \item Warmup ratio: 0.1
    \item Learning rate: 5e-7
    \item Max. seq. length: 8,192
    \item Effective batch size: 32
    \item Beta: 0.1
\end{itemize}

All models except QLoRA models were trained on a 256-chip (512-chip for 70B DPO training) TPU v3 pod. Our training code is based off EasyLM~\citep{geng2023easylm} and available at \href{https://github.com/hamishivi/EasyLM}{https://github.com/hamishivi/EasyLM}.

QLoRA models were trained on an internal A100 80GB cluster using finetuning code available at \href{https://github.com/allenai/open-instruct}{https://github.com/allenai/open-instruct}.

\section{Science Mixture Breakdown}
\label{app:science_mix}
We provide a breakdown of what tasks are included, and their dataset of origin, in our science mixture in Table~\ref{tab:science_mix}.

\input{tables/science_doc_tasks}

\section{Full MT-Bench Results}
\label{app:full_mt_bench}
In Table~\ref{tab:full_mtbench} we show full MT-Bench results, split by category, for all models shown in Table~\ref{tab:mtbench}. We use GPT-4-0613 as the judge model.

\input{tables/full_mtbench}

\end{document}

%% file: tables/baseline_comparisons.tex
\begin{table}[htbp]
\setlength\tabcolsep{2pt}
\centering
\resizebox{\textwidth}{!}{
\begin{tabular}{lcccccccc}
\toprule
{} & \textbf{MMLU}  & \textbf{GSM8k} & \textbf{BBH}  & \textbf{TydiQA GP} & \textbf{CodexEval} & \textbf{AlpacaEval} & \textbf{ToxiGen} & \textbf{Average} \\
{} & \textbf{0-shot, EM} & \textbf{8-shot CoT, EM} & \textbf{3-shot CoT, EM} & \textbf{1-shot, F1} & \textbf{P@10} & \textbf{\% Win} & \textbf{\% Toxic} & - \\
\midrule
\multicolumn{9}{c}{Proprietary models} \\ \midrule
GPT-4-0613 & \textbf{81.4} & \textbf{95.0} & \textbf{89.1} & \textbf{65.2} & 87.0 & 91.2 & 0.6 & \textbf{86.9} \\ 
GPT-3.5-turbo-0613 & 65.7 & 76.5 & 70.8 &  51.2 & 88.0 &  \textbf{91.8} &  \textbf{0.5} & 77.6 \\
GPT-3.5-turbo-0301 & 67.9 & 76.0 & 66.1 &  51.9 & \textbf{88.4} &  83.6 &  27.7 & 72.3 \\
\midrule
\multicolumn{9}{c}{Non-\modelname~Open Models} \\ \midrule
Zephyr-Beta 7B & 58.6 & 28.0 & 44.9 & 23.7 & 54.3 & 86.3 & 64.0 & 47.4 \\
Xwin-LM v0.1 70B & \textbf{65.0} & \textbf{65.5} & \textbf{65.6} & 38.2 & \textbf{66.1} & \textbf{\underline{95.8}} & 12.7 & \textbf{69.1} \\
\llama-2-Chat 7B & 46.8 & 12.0 & 25.6 & 22.7 & 24.0 & 87.3 & \textbf{\underline{0.0}} & 45.4 \\
\llama-2-Chat 13B & 53.2 & 9.0 & 40.3 & 32.1 & 33.1 & 91.4 & \textbf{\underline{0.0}} & 51.3 \\
\llama-2-Chat 70B & 60.9 & 59.0 & 49.0 & \textbf{44.4} & 52.1 & 94.5 & \textbf{\underline{0.0}} & 65.7 \\ \midrule
\multicolumn{9}{c}{\modelname~2 Suite} \\ \midrule
\modelname~2 7B & 50.4 & 34.0 & 48.5 & 46.4 & 36.9 & 73.9 & 7.0 & 54.7 \\
\modelname~2+DPO 7B & 50.7 & 34.5 & 45.5 & 44.5 & 40.0 & 85.1 & 0.5 & 56.3 \\
\modelname~2 13B & 55.4 & 46.0 & 49.5 & 53.2 & 49.0 & 78.9 & 1.7 & 61.5 \\
\modelname~2+DPO 13B & 55.3 & 49.5 & 49.4 & 39.7 & 48.9 & 89.5 & 1.1 & 61.6 \\
\modelname~2 70B & 67.3 & \textbf{\underline{73.0}} & \textbf{\underline{68.4}} & \textbf{\underline{53.6}} & 68.5 & 86.6 & 0.5 & \textbf{\underline{73.8}} \\
\modelname~2+DPO 70B & \textbf{\underline{67.8}} & 71.5 & 66.0 & 35.8 & \textbf{\underline{68.9}} & \textbf{95.1} & \textbf{0.2} & 72.1 \\
\bottomrule
\end{tabular}
} \vspace{4pt}
\caption{
The evaluation metrics of our core \modelname{}-2 suite and its peers.
Most of the models included use \llama~2~base models, except Zephyr-Beta, which uses \mistral-7B. 
For all evaluations except ToxiGen, higher scores are better. We average scores naively, apart from Toxigen, where we take 100 - $x$ as the value to average.
The top-performing open model per task has been underlined, and the top-performing model in each set of models is bolded.
}
\label{tab:baseline_comparisons}
\end{table}

%% file: tables/mix_comparison_table.tex
\begin{table}[]
\setlength\tabcolsep{2pt}
\resizebox{\textwidth}{!}{
\begin{tabular}{ccccccccccc}
\midrule
\textbf{Size} & \textbf{Data} & \textbf{MMLU}   & \textbf{GSM8k}      & \textbf{BBH} & \textbf{TydiQA} & \textbf{Codex-Eval} & \textbf{AlpacaEval} & \textbf{ToxiGen} & \textbf{TruthfulQA} & \textbf{Average}     \\
 & & \textbf{0-shot} & \textbf{8-shot CoT} & \textbf{3-shot CoT} & \textbf{1-shot} & \textbf{Pass@10}    & \textbf{\%win}& \textbf{\% Toxic} & \textbf{\%Info+True} & - \\ \midrule
\multirow{3}{*}{7B} & ShareGPT & 47.8 & 20.0 & 41.5 & 24.0 & 29.2 & 72.3 & 12.6 & \textbf{54.1} & 47.0 \\
& V1 mix.  & 49.2 & \textbf{37.0} & 44.2     & \textbf{52.9}   & 33.9     & 64.5     & 39.9  & 40.8     & 47.8      \\
& V2 mix. & \textbf{50.4} & 34.0 & \textbf{48.5} & 46.4 & \textbf{36.9} & \textbf{73.9} & \textbf{7.0} & 50.2 & \textbf{54.2} \\ \midrule
\multirow{2}{*}{13B} & V1 mix. & 52.3 & \textbf{53.0} & \textbf{50.6} & \textbf{58.8}   & 38.9     & 67.7     & 18.7  & 45.3     & 56.0      \\
& V2 mix. & \textbf{55.4} & 46.0 & 49.5 & 53.2 & \textbf{49.0} & \textbf{78.9} & \textbf{1.7} & \textbf{55.8} & \textbf{60.8} \\ \midrule
\multirow{2}{*}{70B} & V1 mix. & \textbf{67.3}   & \textbf{74.5}& 67.5     & \textbf{56.8}   & 65.4     & 82.8     & \textbf{0.0}     & 57.9     & 71.5      \\
& V2 mix. & \textbf{67.3}   & 73.0 & \textbf{68.4} & 53.6 & \textbf{68.5}& \textbf{86.6} & 0.5   & \textbf{62.2} & \textbf{72.4} \\ \bottomrule
\end{tabular}}
\vspace{4pt}
\caption{Results of \llama-2 models finetuned on our V1 and V2 data mixtures, and ShareGPT.}
\label{tab:mix_comparisons}
\end{table}

%% file: tables/dpo_comparisons.tex
\begin{table}[htbp]
\setlength\tabcolsep{2pt}
\centering

\resizebox{\textwidth}{!}{
\begin{tabular}{@{}clcccccccc@{}}
\toprule
     \textbf{Size} & \textbf{Model} & \textbf{MMLU}   & \textbf{GSM8k}      & \textbf{BBH}        & \textbf{TydiQA} & \textbf{Codex-Eval} & \textbf{AlpacaEval} & \textbf{ToxiGen} & \textbf{Average}     \\
      && \textbf{0-shot} & \textbf{8-shot CoT} & \textbf{3-shot CoT} & \textbf{1-shot} & \textbf{Pass@10}    & \textbf{\%win}       & \textbf{\% Toxic} & \multicolumn{1}{l}{} \\ \midrule
\multirow{3}{*}{7B} &\modelname~2 & 50.4 & 34.0 & \textbf{48.5} & \textbf{46.4} & 36.9 & 73.9 & 7.0 & 54.7 \\
&\modelname~2+DPO & \textbf{50.7} & \textbf{34.5} & 45.5 & 44.5 & \textbf{40.0} & \textbf{85.1} & \textbf{0.5} & \textbf{56.3} \\ \cdashline{2-10}\rule{0pt}{9pt}
&$\Delta$ & \textcolor{darkblue}{+0.3} & \textcolor{darkblue}{+0.5} & \textcolor{darkred}{-3.0} & \textcolor{darkred}{-1.9} & \textcolor{darkblue}{+3.1}     & \textcolor{darkblue}{+11.2}     & \textcolor{darkblue}{-6.5} & \textcolor{darkblue}{+1.6}      \\ \midrule
\multirow{3}{*}{13B} & \modelname~2 & \textbf{55.4} & 46.0 & \textbf{49.5} & \textbf{53.2} & \textbf{49.0} & 78.9 & 1.7 & 61.5 \\
&\modelname~2+DPO & 55.3 & \textbf{49.5} & 49.4 & 39.7 & 48.9 & \textbf{89.5} & \textbf{1.1} & \textbf{61.6} \\ \cdashline{2-10}\rule{0pt}{9pt}
& $\Delta$ & \textcolor{darkred}{-0.1} & \textcolor{darkblue}{+3.5}     & \textcolor{darkred}{-0.1}    & \textcolor{darkred}{-13.5}           & \textcolor{darkred}{-0.1}    & \textcolor{darkblue}{+10.6}     & \textcolor{darkblue}{-0.6} & \textcolor{darkblue}{+0.1}     \\ \midrule
\multirow{3}{*}{70B} & \modelname~2 & 67.3 & \textbf{73.0} & \textbf{68.4} & \textbf{53.6} & 68.5 & 86.6 & 0.5 & \textbf{73.8} \\
&\modelname~2+DPO & \textbf{67.8} & 71.5 & 66.0 & 35.8 & \textbf{68.9} & \textbf{95.1} & \textbf{0.2} & 72.1 \\ \cdashline{2-10}\rule{0pt}{9pt}
& $\Delta$ & \textcolor{darkblue}{+0.5} & \textcolor{darkred}{-1.5}    & \textcolor{darkred}{-2.4}    & \textcolor{darkred}{-17.8}           & \textcolor{darkblue}{+0.4}     & \textcolor{darkblue}{+8.5}     & \textcolor{darkblue}{-0.3} & \textcolor{darkred}{-1.7}     \\ \bottomrule
\end{tabular}}
\vspace{4pt}
\caption{
Evaluation results for \modelname{}~V2 models with and without DPO finetuning, and the difference between the two results ($\Delta$).
}
\label{tab:dpo_comparisons}
\end{table}

%% file: tables/mtbench.tex
\begin{table}
\centering
\setlength\tabcolsep{4pt}
\begin{tabular}{clccc}
\toprule
 \textbf{Size} & \textbf{Model} & \multicolumn{1}{c}{\textbf{MT-Bench}} & \multicolumn{2}{c}{\textbf{AlpacaEval}}    \\
 && \multicolumn{1}{c}{\textbf{Average Score}}      & \textbf{Winrate (\%)} & \textbf{\begin{tabular}[c]{@{}c@{}} Avg. Output\\Length\end{tabular}} \\ \midrule
\multirow{4}{*}{unk.}& GPT-4-1106-preview & \textbf{9.26} & \textbf{97.1} & 2041 \\
&GPT-4-0613                & 9.18               & 91.2              & 1090                  \\
&GPT-3.5-turbo-0613         &  8.39              &  91.8              & 1416                   \\
&GPT-3.5-turbo-0301         & 7.94               &  83.6              & 838                   \\ \midrule
\multirow{3}{*}{7B} & Zephyr-Beta          & \textbf{7.35}               & \textbf{86.3}              & 2721                  \\
&\modelname~2              & 6.30                                      & 73.9              & 1248                  \\
&\modelname~2+DPO          & 6.27                                      & 85.1              & 1437                  \\ \midrule
\multirow{3}{*}{13B} & Xwin v0.2 & \textbf{7.01} & \textbf{91.0}	& 2748 \\
&\modelname~2              & 6.70                                      & 78.9              & 1034                  \\
&\modelname~2+DPO          & 7.00                                      & 89.5              & 1414                  \\ \midrule
\multirow{3}{*}{70B}& Xwin v0.1            & 7.53                                      & \textbf{95.8}              & 1797                  \\
&\modelname~2              & 7.49                                      & 86.6              & 1011                  \\
&\modelname~2+DPO        & \textbf{7.89}                                      & 95.1              & 1414                  \\\bottomrule
\end{tabular}
\vspace{4pt} 
\caption{MT-Bench and AlpacaEval results, along with average output length of AlpacaEval responses. GPT model size is unknown. We include output length to observe the effect of DPO on model verbosity. `GPT-4-1106-preview' is also known as `GPT-4 Turbo' (See \url{https://help.openai.com/en/articles/8555510-gpt-4-turbo}).}
\label{tab:mtbench}
\end{table}

%% file: tables/qlora_comparisons.tex
\begin{table}[]
\setlength\tabcolsep{2pt}
\resizebox{\textwidth}{!}{
\begin{tabular}{@{}clccccccccc@{}}
\toprule
   \textbf{Size} & \textbf{Model} & \textbf{MMLU}   & \textbf{GSM8k} & \textbf{BBH} & \textbf{TydiQA} & \textbf{Codex-Eval} & \textbf{AlpacaEval}  & \textbf{ToxiGen} & \textbf{TruthfulQA} & \textbf{Average} \\
   && \textbf{0-shot} & \textbf{8-shot CoT} & \textbf{3-shot CoT} & \textbf{1-shot} & \textbf{Pass@10} & \textbf{\%win} & \textbf{\% Toxic} & \textbf{\%Info+True} & \multicolumn{1}{l}{} \\ \midrule
\multirow{3}{*}{7B} & \llama{}-2 base & 41.8  & 12.0 & 39.3 & \textbf{51.2}  & 26.8 & - & 77.3   & 26.7 & - \\
&\modelname{}~2  & \textbf{50.4}   & \textbf{34.0} & \textbf{48.5} & 46.4  & \textbf{36.9} & \textbf{73.9} & \textbf{7.0} & 40.8 & \textbf{53.0} \\
&\modelname{}~2 (QLoRA)  & 48.8  & 20.5 & 45.7 & 49.2   & 31.7 & 56.1 & 14.7  & \textbf{44.6} & 47.7 \\\midrule
\multirow{3}{*}{13B} &\llama{}-2 base & 52.0  & 25.0 & 48.9 & \textbf{56.5}  & 32.5 & - & 85.7   & 31.1 & - \\
&\modelname{}~2  & \textbf{55.4}   & \textbf{46.0} & 49.5 & 53.2  & \textbf{49.0} & \textbf{78.9} & 1.7 & \textbf{55.8} & \textbf{60.8} \\
&\modelname{}~2 (QLoRA) & 54.6  & 36.0 & \textbf{52.5} & 54.6   & 39.1 & 65.6 & \textbf{0.0} & 55.2 & 57.2 \\\midrule
\multirow{3}{*}{70B} & \llama{}-2 base & 64.5  & 55.5 & 66.0 & \textbf{62.6}  & 60.1 & - & 84.2   & 38.2 & - \\
&\modelname{}~2  & 67.3  & \textbf{73.0} & 68.4 & 53.6  & \textbf{68.5} & \textbf{86.6} & \textbf{0.5} & \textbf{62.2} & \textbf{73.4} \\
&\modelname{}~2 (QLoRA) & \textbf{67.4}   & 64.5 & \textbf{71.6} & 60.9   & 66.9 & 78.6 & \textbf{0.5} & 58.4 & 71.0 \\\bottomrule 
\end{tabular}}
\vspace{4pt}
\caption{Results from \llama-2 models finetuned with and without QLoRA on our V2 mix. 
We also report results from \llama-2 models without any finetuning (base).}
\label{tab:qlora_comparisons}
\end{table}

%% file: tables/codellama_comparisons.tex
\begin{table}[htbp]
\setlength\tabcolsep{2pt}
\centering
\resizebox{\textwidth}{!}{
\begin{tabular}{@{}clccccccccc@{}}
\toprule
    \textbf{Size} & \textbf{Model} & \textbf{MMLU}   & \textbf{GSM8k}  & \textbf{BBH}    & \textbf{TydiQA} & \textbf{Codex-Eval} & \textbf{AlpacaEval}  & \textbf{ToxiGen} & \textbf{TruthfulQA} & \textbf{Average} \\
    && \textbf{0-shot} & \textbf{8-shot CoT} & \textbf{3-shot CoT} & \textbf{1-shot} & \textbf{Pass@10}    & \textbf{\%win}    & \textbf{\% Toxic} & \textbf{\%Info+True} &  \\ \midrule
\multirow{3}{*}{7B} & \codellama{} base  & 33.8 & 12.0 & 43.4 & 47.6 & 58.7 & - & 81.5  & 26.1 & -  \\
&\codellama{}~Instruct & 41.5 & 17.0 & 38.4 & 41.6 & 64.1 & 71.9  & \textbf{1.0} & 15.2 & 48.6  \\
&\modelname~2   & \textbf{50.4}   & \textbf{34.0} & 48.5 & 46.4  & 36.9 & \textbf{73.9} & 7.0 & \textbf{40.8} & 53.0 \\
&\codemodelname{}~2 & 43.7 & 33.0 & \textbf{49.1}   & \textbf{52.6}   & \textbf{68.9}   & 58.0  & 5.0   & 33.0 & \textbf{54.2}    \\ \midrule
\multirow{3}{*}{13B} & \codellama{} base & 37.5 & 22.0 & 49.5 & 52.1 & 69.8 & - & 77.9  & 26.9 &  - \\
&\codellama{}~Instruct & 43.3 & 23.0 & 48.0 & 37.8 & 69.2 & 75.3  & \textbf{0.0}   & 38.1 & 54.3  \\
&\modelname~2  & \textbf{55.4}   & \textbf{46.0} & 49.5 & 53.2  & 49.0 & \textbf{78.9} & 1.7 & \textbf{55.8} & \textbf{60.8} \\
&\codemodelname{}~2    & 45.9 & 41.0 & \textbf{52.8}   & \textbf{55.7}   & \textbf{76.2}   & 64.1   & \textbf{0.0}   & 36.7 & 59.1  \\ \midrule
\multirow{3}{*}{34B} & \codellama{} base & 47.4 & 35.0 & 57.0 & 57.1 & 77.6 & - & 88.3  & 24.4 & -  \\
&\codellama{}~Instruct & 50.9 & 38.0 & 59.2 & 55.1 & 76.5 & \textbf{84.5} & \textbf{0.0} & \textbf{51.2}   & 64.4 \\
&\codemodelname{}~2    & \textbf{53.6}   & \textbf{54.0}   & \textbf{64.3}   & \textbf{60.6}   & \textbf{82.5}   & 76.8  & \textbf{0.0} & 42.0 & \textbf{66.7}    \\\bottomrule
\end{tabular}}
\vspace{4pt}
\caption{
Evaluation results comparing models based on \codellama{}~with our \modelname{} models.
\codemodelname{}~2 refers to \codellama{}~models finetuned on our V2 mixture.}
\label{tab:codellama_comparisons}
\end{table}

%% file: tables/science_doc_tasks.tex
\begin{table}[htbp]
    \setlength\tabcolsep{2pt}
    \centering
    \resizebox{\textwidth}{!}{
        \begin{tabular}{lll}
            \toprule
            \textbf{Dataset}                                      & \textbf{Tasks}                                                        & \textbf{\# Examples} \\
            \midrule
            Evidence Inference \citep{lehman-etal-2019-inferring} & Information extraction: Medical evidence 5-tuples                     & 1,678                \\
            Qasper             \citep{dasigi-etal-2021-dataset}   & Question answering                                                    & 2,255                \\
            SciERC             \citep{luan-etal-2018-multi}       & Information extraction: Named entity recognition, Relation extraction & 700                  \\
            SciFact            \citep{wadden-etal-2020-fact}      & Fact checking                                                         & 919                  \\
            SciTLDR            \citep{cachola-etal-2020-tldr}     & Summarization                                                         & 1,992                \\
            \bottomrule
        \end{tabular}
    }
    \vspace{4pt}
    \caption{Datasets included in the science literature instruction mix for \modelname{} V2.}
    \label{tab:science_mix}
\end{table}

%% file: tables/full_mtbench.tex
\begin{table}[h]
\setlength\tabcolsep{5pt}
\resizebox{\textwidth}{!}{
\begin{tabular}{@{}lccccccccc@{}}
\toprule
                & \multicolumn{1}{l}{\textbf{STEM}} & \multicolumn{1}{l}{\textbf{Humanities}} & \multicolumn{1}{l}{\textbf{Reasoning}} & \multicolumn{1}{l}{\textbf{Coding}} & \multicolumn{1}{l}{\textbf{Math}} & \multicolumn{1}{l}{\textbf{Extraction}} & \multicolumn{1}{l}{\textbf{Roleplay}} & \multicolumn{1}{l}{\textbf{Writing}} & \multicolumn{1}{l}{\textbf{Average}} \\ \midrule
\multicolumn{10}{c}{Proprietary Models}                                                                                                                                                                                                                                                                                                                                          \\ \midrule
GPT-4-1106-preview     & \textbf{9.90} & \textbf{9.95} & 8.10 & \textbf{9.05} & 7.95 & \textbf{9.90} & \textbf{9.50} & \textbf{9.70} & \textbf{9.26}  \\
GPT-4-0613 & 9.65 & 9.85 & \textbf{9.30} & 8.60 & \textbf{8.10} & 9.35 & 9.03 & 9.55 & 9.18 \\
GPT-3.5-turbo-0613 & 9.55 & \textbf{9.95} & 6.20 & 7.05 & 7.05 & 9.00 & 8.65 & 9.65 & 8.39 \\
GPT-3.5-turbo-0301     & 9.05 & 9.55 & 6.30 & 6.70 & 5.20 & 8.60 & 8.55	& 9.60 & 7.94                                 \\\midrule
\multicolumn{10}{c}{Open Models}                                                                                                                                                                                                                                                                                                                                                 \\ \midrule
\llama{}-2-Chat 7B  & 8.65                              & 8.75                                    & 4.25                                   & 3.00                                & 2.40                              & 6.50                                    & 7.70                                  & 8.90                                 & 6.27                                 \\
\llama{}-2-Chat 13B & 8.63                              & 9.75                                    & 5.10                                   & 3.00                                & 3.45                              & 6.93                                    & 7.50                                  & 8.85                                 & 6.65                                 \\
\llama{}-2-Chat 70B & 8.93                              & 9.63                                    & 5.80                                   & 3.15                                & 3.30                     & 7.25                                    & 7.50                                  & 9.30                                 & 6.86                                 \\
Zephyr-Beta 7B & 9.03 & 9.63 & 5.60 & \textbf{\underline{5.10}} & \textbf{4.45} & 7.45 & 8.20 & 9.35 & 7.35 \\
Xwin 70b v0.1   & \textbf{\underline{9.68}}                     & \textbf{\underline{9.95}}                           & \textbf{6.55}                          & 4.25                       & 3.30                    & \textbf{8.75}                           & 8.25                                  & \textbf{\underline{9.55}}                        & \textbf{7.53}                        \\
Xwin 13b v0.2   & 9.55                              & 9.88                                    & 5.20                                   & 3.60                                & 2.85                              & 7.70                                    & \textbf{8.60}                         & 8.68                                 & 7.01                                 \\ \midrule
\multicolumn{10}{c}{\modelname{} V2 Models}                                                                                                                                                                                                                                                                                                                                              \\ \midrule
\modelname~2 7B      & 8.00                              & 9.50                                    & 4.40                                   & 3.40                                & 3.30                              & 6.10                                    & 7.63                                  & 8.10                                 & 6.30                                 \\
\modelname~2+DPO 7B  & 8.23                              & 9.60                                    & 4.30                                   & 3.32                                & 2.35                              & 6.05                                    & 7.95                                  & 8.35                                 & 6.27                                 \\
\modelname~2 13B     & 8.70                              & 9.25                                    & 5.45                                   & 4.30                                & 3.75                              & 7.35                                    & 7.50                                  & 7.30                                 & 6.70                                 \\
\modelname~2+DPO 13B & \textbf{9.08}                     & 9.80                                    & 5.30                                   & 3.60                                & 2.95                              & 8.00                                    & 8.60                                  & 8.70                                 & 7.00                                 \\
\modelname~2 70B     & 9.00                              & 9.75                                    & 5.50                                   & \textbf{\underline{5.10}}                       & \textbf{\underline{4.70}}                     & 8.45                                    & 8.30                                  & 9.15                                 & 7.49                                 \\
\modelname~2+DPO 70B & 9.00                              & \textbf{9.90}                           & \textbf{\underline{7.00}}                          & 4.70                                & 4.65                              & \textbf{\underline{9.35}}                           & \textbf{\underline{9.25}}                         & \textbf{9.25}                        & \textbf{\underline{7.89}}                        \\ \bottomrule
\end{tabular}
}
\vspace{4pt}
\caption{Full MT-Bench results split by category. Score is an average of scores given by a GPT-4 annotator. The best open-weight model performance is underlined.}
\label{tab:full_mtbench}
\end{table}